\documentclass[letterpaper, 10 pt, journal]{IEEEtran}
\IEEEoverridecommandlockouts                          
\usepackage{graphics} 
\usepackage{epsfig}
\usepackage{times}
\usepackage{amsmath}
\usepackage{amssymb}
\usepackage{threeparttable}
\usepackage{booktabs}
\usepackage{soul}
\usepackage{subfigure}
\usepackage{xcolor}
\usepackage[hidelinks]{hyperref}

\title{\LARGE \bf
Bayesian-Optimized One-Step Diffusion Model with Knowledge Distillation for Real-Time 3D Human Motion Prediction
}

\author{Sibo Tian$^{1}$, Minghui Zheng$^{1,*}$, and Xiao Liang$^{2,*}$
\thanks{This work was supported by the USA National Science Foundation under Grant No. 2026533/2422826. The authors confirm that all human/animal subject research procedures and protocols are approved by the review board. Portions of this research were conducted with the advanced computing resources provided by Texas A\&M High Performance Research Computing.}
\thanks{$^{1}$ Sibo Tian and Minghui Zheng are with the J. Mike Walker '66 Department of Mechanical Engineering, Texas A\&M University, College Station, TX 77843, USA. {\tt\small Emails: {sibotian, mhzheng}@tamu.edu.}}
\thanks{$^{2}$ Xiao Liang is with the Zachry Department of Civil and Environmental Engineering, Texas A\&M University, College Station, TX 77843, USA. {\tt\small Email: xliang@tamu.edu.}}
\thanks{$^{*}$ Corresponding Authors.}
}

\begin{document}

\maketitle
\thispagestyle{empty}
\pagestyle{empty}

\begin{abstract}
Human motion prediction is a cornerstone of human-robot collaboration (HRC), as robots need to infer the future movements of human workers based on past motion cues to proactively plan their motion, ensuring safety in close collaboration scenarios. The diffusion model has demonstrated remarkable performance in predicting high-quality motion samples with reasonable diversity, but suffers from a slow generative process which necessitates multiple model evaluations, hindering real-world applications. To enable real-time prediction, in this work, we propose training a one-step multi-layer perceptron-based (MLP-based) diffusion model for motion prediction using knowledge distillation and Bayesian optimization. Our method contains two steps. First, we distill a pretrained diffusion-based motion predictor, \textit{TransFusion}, directly into a one-step diffusion model with the same denoiser architecture. Then, to further reduce the inference time, we remove the computationally expensive components from the original denoiser and use knowledge distillation once again to distill the obtained one-step diffusion model into an even smaller model based solely on MLPs. Bayesian optimization is used to tune the hyperparameters for training the smaller diffusion model. Extensive experimental studies are conducted on benchmark datasets, and our model can significantly improve the inference speed, achieving real-time prediction without noticeable degradation in performance.
\end{abstract}


\section{Introduction}

Human-robot collaboration (HRC) has attracted increasing attention due to its potential in end-of-life product recycling \cite{lee2024review}. When end-of-life products arrive at remanufacturing sites, the considerable uncertainties in both their quantity and quality pose significant challenges to robotic systems that rely on precise and consistent inputs. In such scenarios, human expertise is invaluable for making flexible decisions, and adapting to unexpected conditions that may arise during the disassembly process. Robots, on the other hand, excel at handling hazardous components, heavy lifting, and repetitive tasks. By combining the strength of both humans and robots, HRC can enhance efficiency and safety of product disassembly.

Unlike traditional automation in manufacturing where robots and humans are physically isolated, working in close proximity involves complicated interactions between different agents. In such cases, robots must be capable of understanding human behaviors and predicting human future motion to proactively react to their collaborators, reducing the risk of serious accidents and ensuring safety during collaboration \cite{liu2024integrating, tian2023optimization, zhang2024early, zhang2023unsupervised}. To achieve this, human motion prediction algorithms that gather the real-time data on human movements and use machine learning models to anticipate future actions based on past behaviors must be included in HRC systems.

Despite the good results demonstrated by previous works that forecast human motion in a deterministic way \cite{martinez2017human, cai2020learning}, predicting human motion remains challenging due to the inherently multi-modal nature and associated uncertainties. Human motion is complex and can vary greatly depending on intent, context, and external factors such as interactions with other agents and sudden changes in the environment. Even with similar historical motion, future human movements can differ significantly. These variabilities make it difficult to accurately predict future human motion, leading to limitations in reliability of a single prediction result, especially for safety-critical applications such as HRC. Deep ensembles are used to quantify the prediction uncertainty in \cite{eltouny2024tgn}; however, this work is still based on a deterministic prediction method and cannot handle the diverse patterns of human motion. Rather than relying on a deterministic prediction result, stochastic motion prediction \cite{yan2018mt, yuan2020dlow, gurumurthy2017deligan, barsoum2018hp, wei2023human, barquero2023belfusion, chen2023humanmac, tian2024transfusion}, which is trained to learn the motion distribution and generates multiple predictions, is more effective at ensuring safety in HRC.

\begin{figure}
    \begin{center}
    \includegraphics[width=1.0\columnwidth]{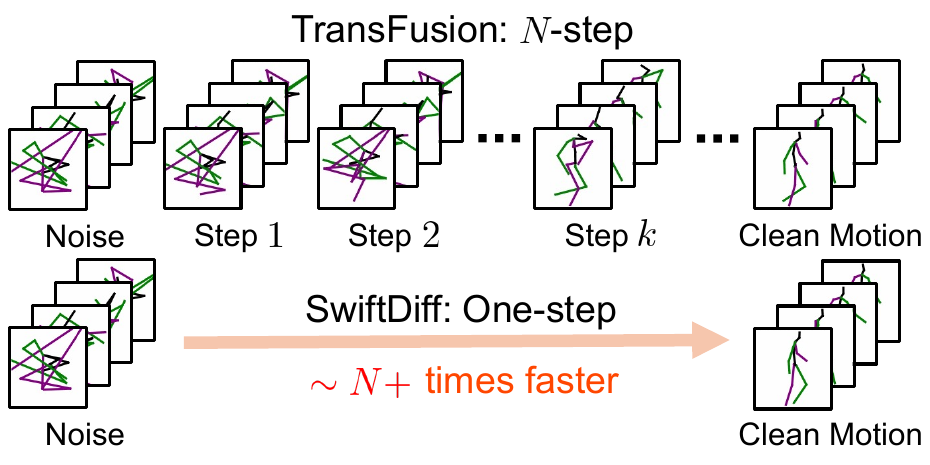}
    \caption{SwiftDiff is a fast one-step MLP-based diffusion model for real-time 3D human motion prediction, derived from TransFusion \cite{tian2024transfusion}. It generates predictions of similar quality much faster than existing diffusion-based methods.}
    \label{overview}
    \end{center}
    \vspace{-0.2in}
\end{figure}

Variational autoencoders (VAEs) \cite{yan2018mt, yuan2020dlow} and generative adversarial networks (GANs) \cite{gurumurthy2017deligan, barsoum2018hp} have been utilized to generate multiple prediction samples given the historical context. Multiple loss terms are involved in these works to explicitly enhance the diversity of prediction results while maintaining good best-of-many \cite{yuan2020dlow} accuracy. Diffusion models \cite{wei2023human, barquero2023belfusion, chen2023humanmac, tian2024transfusion} have received increasing attention recently due to their ability to generate higher-quality prediction samples than the other stochastic motion predictors. While most works focus on achieving greater diversity blindly, often leading to early deviations from the ground truth or sudden stagnation, TransFusion \cite{tian2024transfusion} emphasizes the importance of meaningful diversity and overall prediction quality, as excessive variety and out-of-context predictions may hinder downstream applications such as motion planning of collaborative robots. Although TransFusion showcases its accuracy advantage, it suffers from slow inference speed due to the iterative denoising process. The inference time is critically important in practical applications such as HRC, where robots need to process information and react to their collaborators in real time. A slow prediction speed can cause significant lag, affecting system efficiency and safety.

In this work, we propose training a fast one-step MLP-based diffusion model for real-time 3D human motion prediction using knowledge distillation and Bayesian optimization, as shown in Fig. \ref{overview}. Our method consists of two steps, which helps to gradually reduce the complexity of the distillation task. First, we use a state-of-the-art pretrained diffusion-based human motion prediction model, TransFusion, as the teacher model to distill a one-step diffusion model with the same denoiser architecture. By using the same network structure and initializing the student model with parameters of pretrained model, we can effectively transfer the knowledge and learned features from the teacher model to the student model. Once we have the one-step diffusion model, we then remove the computationally expensive components from the original neural network to create an even smaller and faster denoiser based on MLPs. We use knowledge distillation again, utilizing the one-step predictor obtained in the previous step as the new teacher to train the MLP-based model. Since the structures of the teacher and the student are different this time, we have to randomly initialize the student network's parameters, making the training process more challenging. To achieve good performance, we use Bayesian optimization to search for hyperparameters in the second distillation stage, such as the number of layers, layer dimensions, and learning rate. Overall, the main contributions of this work are summarized as follows:

\begin{itemize}
\item We propose a fast one-step MLP-based diffusion model for human motion prediction. To the best of our knowledge, we are the first to use one-step diffusion to address the real-time human motion prediction problem.
\item We split the distillation problem into two stages, reducing the complexity of the task, and use Bayesian optimization to tune hyperparameters during training.
\item We conduct comprehensive experiments to validate the effectiveness of our method. The results show that our model can significantly improve inference speed, achieving real-time 3D human motion prediction without noticeable degradation in the overall accuracy.
\end{itemize}

\section{Related Work}

\subsection{Diffusion-based Human Motion Prediction}

Diffusion models \cite{ho2020denoising} are a new and promising class of deep generative models that excel in generating high-quality samples through a simple training process. A typical diffusion model consists of two key components: the diffusion process and the denoising process. The diffusion process adds noise to the data following a predefined noise schedule over a series of steps, turning the original data into pure noise. The denoising process employs a neural network to predict the noise injected at each time step, and reverses the diffusion process by gradually removing the noise from the disturbed data. Diffusion models suffer less from the mode collapse problem than other generative models, and considering their ability to generate diverse and high-fidelity samples, diffusion models have been utilized to predict human motion in a stochastic manner based on a short observation.

MotionDiff \cite{wei2023human} utilizes a spatial-temporal transformer as the denoiser in the diffusion model to predict diverse human motion. The denoising process is conditioned on the observation and the denoising step, which are incorporated into the denoiser multiple times within a single denoising step using a linear transformation-based gating and bias mechanism. It then integrates a graph convolutional network (GCN)-based motion refiner to address geometric constraints and enhance prediction quality. BeLFusion \cite{barquero2023belfusion} interprets the diversity of human motion from a behavioral perspective rather than skeleton joint dispersion. The authors train a latent diffusion model, using a conditional U-Net with cross-attention structure \cite{dhariwal2021diffusion} as denoiser. Behavior codes are sampled from latent diffusion model, and then a behavior coupler is utilized to transfer the sampled behavior codes to ongoing motion. Although these two works have demonstrated good performance, they require multiple training stages as they both employ additional networks outside the diffusion model for motion prediction.

End-to-end diffusion-based motion prediction models are proposed in \cite{chen2023humanmac} and \cite{tian2024transfusion}. These two works interpret the motion prediction task as an inpainting problem and solve it in the frequency domain by applying discrete cosine transform (DCT). HumanMAC \cite{chen2023humanmac} uses adaptive normalization modules after each self-attention layer and feed-forward network in the Transformer to guide the prediction with conditions. Unlike previous works that use either cross-attention or adaptive operations to introduce conditions, TransFusion \cite{tian2024transfusion} avoids utilizing additional modules for guiding the denoising process. Instead, it treats both conditions and motion inputs as tokens for the Transformer. To boost the model performance, TransFusion adds lightweight squeeze-and-excitation (SE) blocks \cite{hu2018squeeze} before self-attention layers to dynamically recalibrate all the tokens, giving higher weights to tokens carrying more important information. Though TransFusion greatly simplifies the network structure and achieves state-of-the-art performance in terms of prediction accuracy, it still suffers from slow inference speed, as diffusion models typically require multiple neural network evaluations to generate clean samples from white noise. This hinders the application of diffusion-based motion prediction models in real-world safety-critical cases, such as HRC, where collaborative robots need to react to the anticipated human motion in real-time for the safety and efficiency purpose. To the best of our knowledge, no previous work has explored fast diffusion models for real-time human motion prediction problem.

\begin{figure*}
    \begin{center}
    \includegraphics[width=1.8\columnwidth]{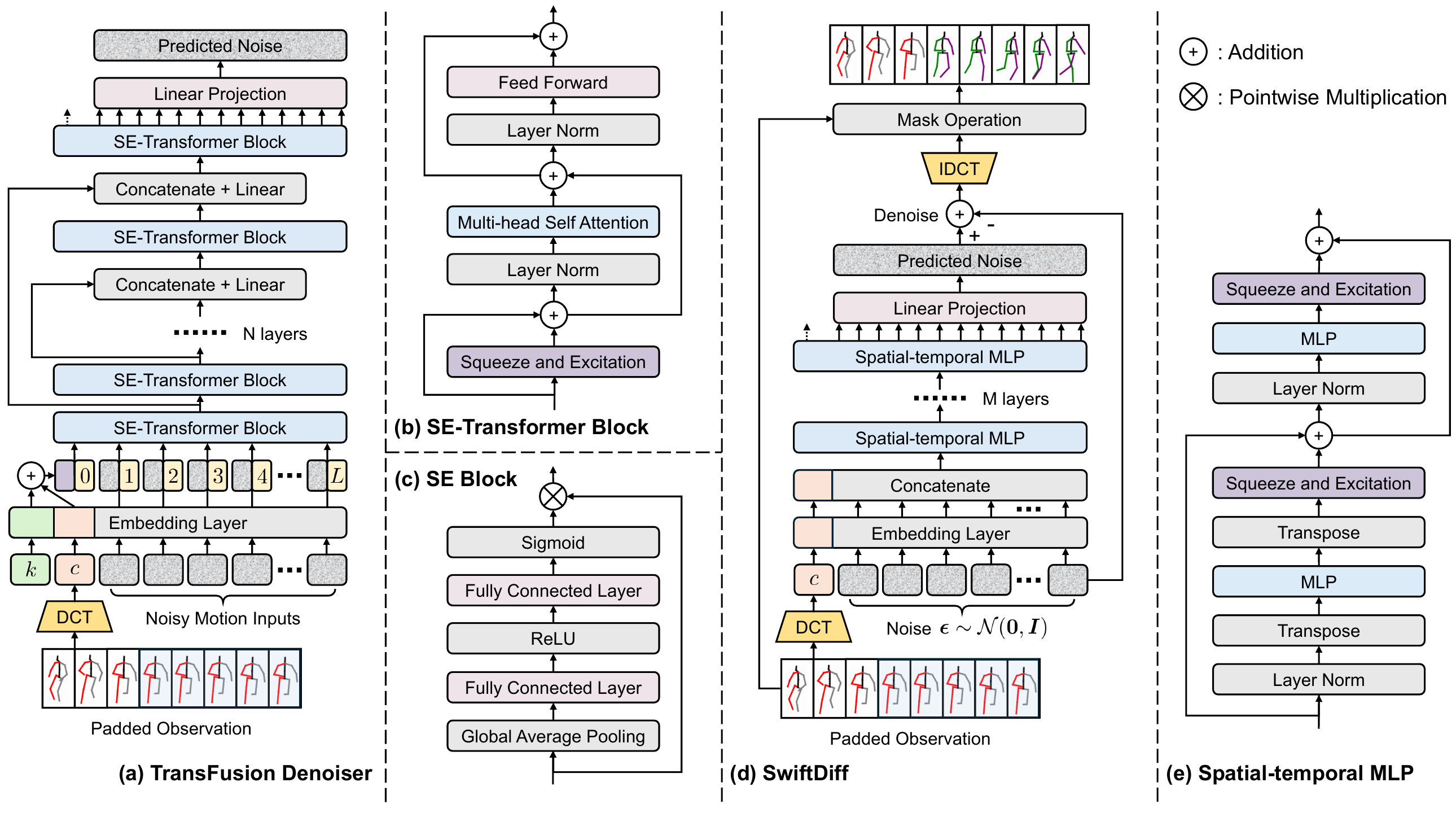}
    \caption{Architecture of the noise prediction network. Figures a, b, and c show the detailed structure of TransFusion, which is used as the teacher model in this work. Figures d and e show the structure of our proposed one-step MLP-based diffusion model for real-time human motion prediction.}
    \label{architecture}
    \end{center}
    \vspace{-0.2in}
\end{figure*}

\subsection{Fast Diffusion Models via Distillation}

Knowledge distillation \cite{hinton2015distilling} transfers the learned features and knowledge of a pretrained teacher model to a student model by minimizing the discrepancies between the outputs of the teacher and the student, and has been applied to diffusion models to significantly accelerate the sampling speed for the image generation task \cite{luhman2021knowledge, salimans2022progressive, song2023consistency, liu2023instaflow}. A straightforward distillation of a pretrained diffusion model is proposed in \cite{luhman2021knowledge}. The authors use Kullback–Leibler (KL) divergence as a distance measure between the outputs of the teacher and the student to train the neural network. Although their method is computationally expensive, as it needs to construct a large dataset of noise-image pair prior to distillation by running the teacher model at its full number of sampling steps, it demonstrates the great potential of training fast diffusion models via knowledge distillation. To avoid running the full number of sampling steps of the original model, a progressive distillation strategy is presented in \cite{salimans2022progressive}. Instead of distilling the original model into a one-step or few-step diffusion model all at once, progressive distillation trains the fast diffusion model iteratively: each time, the algorithm halves the number of required sampling steps of the teacher model, and the resulting student model is used as the new teacher model in the next iteration. With little degradation in sample quality, progressive distillation can produce models that generate high quality images in as few as 4 denoising steps. Observing that, given a pretrained denoising diffusion implicit model (DDIM) \cite{song2021denoising}, trajectories from the sampled noise to the clean data are fixed, the consistency model \cite{song2023consistency} exploits the self-consistency property by training the student model to predict the same outputs when given two adjacent points along the same solution trajectory. In addition to studying the distillation algorithm itself, InstaFlow \cite{liu2023instaflow} introduces a reflow procedure \cite{liu2022flow} prior to distillation, which can significantly enhance the distillation task by straightening the curved trajectories of probability flow ordinary differential equations (ODEs) and refining the coupling between noise and images.

\section{Methodology}

\subsection{Notation}

We use $\boldsymbol{x}$ to represent the sequence of human motion, defined as  $\boldsymbol{x}=\left[\boldsymbol{q}^{(t-H)}, \ldots, \boldsymbol{q}^{(t)}, \ldots , \boldsymbol{q}^{(t+F-1)}\right] \in \mathbb{R}^{(H+F) \times 3J}$, where $\boldsymbol{q}^{(t)} \in \mathbb{R}^{3J}$ denotes the Cartesian coordinates of the human skeleton at time frame $t$, and $J$ is the total number of human joints in the human skeleton. The first $H$ frames of $\boldsymbol{x}$ represent the observation $\boldsymbol{x}^O$, and the subsequent $F$ frames represent the future motion $\boldsymbol{x}^P$ to be predicted given the observation. We use $\boldsymbol{y}$ to represent the frequency components after applying DCT operation to $\boldsymbol{x}$. The compacted observation used as a condition for prediction is represented by $\boldsymbol{c}$. To avoid confusion, we use $t$ to denote time frame and $k$ to indicate the denoising step.

\subsection{TransFusion}

In this paper, we use our prior work, TransFusion \cite{tian2024transfusion}, a state-of-the-art diffusion-based motion predictor, as the teacher model to distill a fast diffusion model for real-time human motion prediction. TransFusion uses Transformer as the denoiser backbone and solves the prediction problem in the frequency domain by applying DCT operation to the motion sequence $\boldsymbol{x}$, i.e., $\boldsymbol{y}=DCT(\boldsymbol{x})$, and its architecture is given in Fig. \ref{architecture}. As the most critical and relevant information of human motion is contained in the lower frequencies, and the higher frequency terms are primarily associated with noise, TransFusion only retains the first $L$ rows of DCT basis to reduce the data dimension. During the diffusion and denoising process, noise is added to and removed from frequency coefficient $\boldsymbol{y}$. The observation $\boldsymbol{x}^O$ is first padded to match the length of the motion sequence, and then we can get the compacted historical motion cues by applying DCT to padded observed sequence, i.e., $\boldsymbol{c} = DCT(Padding(\boldsymbol{x}^{O}))$. Different from other works that use cross-attention or adaptive operations multiple times within a single denoising step to guide the prediction with conditions including the observation and denoising step, TransFusion simply treats the conditions and noisy motion frequency coefficients as tokens and lets the neural network learn the interconnections among them.

TransFusion consists of several SE-Transformer layers with long skip connections between shallow and deep layers. Skip connections are achieved by first concatenating tensors from two branches and then passing them through a linear layer to match the dimensions. SE blocks \cite{hu2018squeeze}, which are a type of lightweight attention mechanism consisting of only two fully connected layers and a single pointwise multiplication, are added before each self-attention layer to adaptively rescale all the tokens, optimizing the Transformer's learning process and thus enhancing network performance.

TransFusion treats the motion prediction task as an inpainting problem. In other words, for a complete motion sequence, the task of human motion prediction is equivalent to reconstructing the missing parts given the first few elements of the motion sequence. During inference, for each denoising step $k$, we need to replace the first $H$ frames of the human motion sequence with the noisy ground truth by doing the following masking operation, before finishing the current denoising step and starting the next round of denoising:
$$\boldsymbol{x}_{k}= \boldsymbol{M} \odot IDCT(\boldsymbol{y}_{k}^{O}) + (\boldsymbol{1}-\boldsymbol{M}) \odot IDCT(\boldsymbol{y}_{k}^{D} ),  \eqno{(1)} $$
$$\boldsymbol{y}_{k} = DCT(\boldsymbol{x}_{k}),  \eqno{(2)}$$
where $\boldsymbol{M}$ is the mask defined as $\boldsymbol{M}=[1, \ldots, 1,0, \ldots 0]^{\top}$ with the first $H$ elements equal to 1, $\boldsymbol{y}_{k}$ is the final noisy frequency coefficients obtained from denoising step $k$ and also serves as the input for the next denoising step, $\boldsymbol{y}_{k}^{O}$ is the noisy sample obtained by adding noise to $\boldsymbol{c}$ according to the diffusion noise schedule, and $\boldsymbol{y}_{k}^{D}$ is the sample derived by subtracting the predicted noise from $\boldsymbol{y}_{k+1}$. Detailed information on the neural network structure, training, and inference process of TransFusion can be found in \cite{tian2024transfusion}.

\subsection{Knowledge Distillation}

\begin{figure}
    \begin{center}
    \includegraphics[width=1.0\columnwidth]{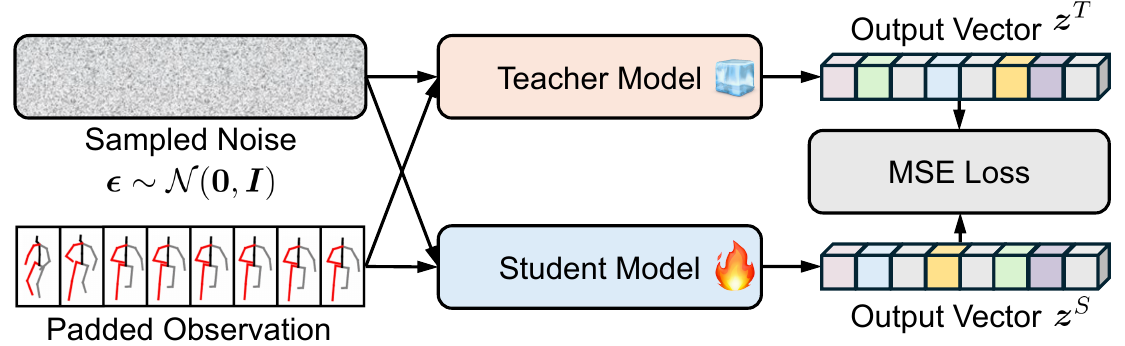}
    \caption{Overview of knowledge distillation with mean squared error loss. The parameters of teacher model are frozen during distillation. only the parameters of student model are updated.}
    \label{distillation}
    \end{center}
    \vspace{-0.2in}
\end{figure}

Since the motion prediction task is strongly conditioned on the observation, which allows the denoising process to be guided more effectively, sampling can be completed in relatively fewer steps during the denoising stage compared to tasks with less conditioning information. For example, TransFusion is trained with one thousand diffusion steps but can achieve state-of-the-art performance in as few as 20 denoising steps. In this case, a straightforward knowledge distillation method as shown in Fig. \ref{distillation}, which directly and explicitly supervises the discrepancies between the outputs of the teacher and the student, can be utilized in this work. Unlike \cite{luhman2021knowledge}, which uses KL divergence as the loss function, we use mean squared error (MSE) as the discrepancy measure to train the knowledge distillation since it shows better results compared with KL divergence loss in logit matching tasks \cite{kim2021comparing}. The loss function for knowledge distillation in this work is formatted as follows:
$$\mathcal{L} = \mathbb{E}_{\boldsymbol{x}^O, \boldsymbol{\epsilon}} \left[\left\| \mathcal{F}_{teacher}(\boldsymbol{x}^O, \boldsymbol{\epsilon}) - \mathcal{F}_{student}(\boldsymbol{x}^O, \boldsymbol{\epsilon})\right\|_2^2\right],  \eqno{(3)} $$
where $\boldsymbol{x}^O$ is the observed motion sampled from dataset, $\boldsymbol{\epsilon} \sim \mathcal{N}(\boldsymbol{0}, \boldsymbol{I})$ is the noise sampled from the standard normal distribution, $\mathcal{F}_{teacher}(\boldsymbol{x}^O, \boldsymbol{\epsilon})$ and $\mathcal{F}_{student}(\boldsymbol{x}^O, \boldsymbol{\epsilon})$ are the outputs from the teacher the student models, respectively.

Our strategy for distilling TransFusion into a smaller model for real-time human motion prediction consists of two stages. In the first step, we use pretrained TransFusion as the teacher model to distill a one-step diffusion model with an identical neural network structure. We initialize the student model with parameters of the well-trained TransFusion to effectively transfer the knowledge from teacher to student, leading to faster convergence. 

In the second phase, to further accelerate the inference process and reduce the size of the neural network, we replace TransFusion, which includes several computationally expensive and parameter-heavy components, such as self-attention modules and concatenation-based long-skip connections, with a slimmed-down denoising neural network based entirely on MLPs. We then use the one-step TransFusion model obtained in the first distillation stage as the new teacher model to distill this one-step MLP-based student diffusion model, which is initialized randomly.

This two-stage distillation strategy offers several benefits. First, it breaks the distillation process into two parts, gradually reducing the complexity of the task and preserving the learned knowledge more effectively than directly distilling the pretrained TransFusion model into a one-step diffusion model with a much simpler denoiser. Additionally, it helps reduce the effort required to tune the MLP-based denoiser, since the one-step TransFusion model operates much faster than the original model, resulting in less training time.

\subsection{SwiftDiff}

We name the proposed one-step MLP-based diffusion model for real-time motion prediction as SwiftDiff and its structure is shown in Fig. \ref{architecture}. The denoiser of SwiftDiff consists of three main components: layer normalization, MLP blocks, and SE blocks. Since there is no self-attention layer to learn the relationships between tokens, MLP blocks are applied to both the sequence dimension and the channel dimension to mix the information. Similar to TransFusion, we first apply DCT to the padded observed motion to obtain compact history motion cues. Then, this condition and noise sample inputs are passed to embedding layers to perform channel lifting. After that, the embedded observation and noise samples are concatenated along the sequence dimension and passed through the model to predict the noise. Since SwiftDiff is a one-step diffusion model, the noise prediction does not need to conditioned on the diffusion step. Finally, the pure motion sequence in the time domain is obtained by applying IDCT to the denoised frequency components.

\subsection{Bayesian Optimization}

During the parameter tuning process of SwiftDiff, we aim to find an optimal set of model hyperparameters, denoted as $\boldsymbol{\lambda}$, that minimizes a customized performance metric $\boldsymbol{g}$, which can be the MSE between the outputs of the teacher and the student, the model inference time, the prediction error compared with the human motion ground truth, or a weighted combination of these factors. Such optimization problem can be summarized as:
$$\min_{\boldsymbol{\lambda}} \, \mathcal{G}(\boldsymbol{\lambda}) \quad \text{subject to} \quad \boldsymbol{\lambda} \in \boldsymbol{\Lambda}. \eqno{(4)}$$
This objective function $\mathcal{G}(\boldsymbol{\lambda})$ is evidently non-convex and lacks a closed-form expression, meaning we can only obtain observations of $\boldsymbol{g}=\mathcal{G}(\boldsymbol{\lambda})$ at certain sampled values of $\boldsymbol{\lambda}$. A common practice is to perform a grid search over the possible parameter space $\boldsymbol{\Lambda}$ and choose the model hyperparameters with the best evaluation metric. In this work, we employ a more appealing approach, Bayesian optimization, which has been generally proven to be effective in finding the minimum of complex non-convex functions with relatively few evaluations by balancing exploration and exploitation \cite{brochu2010tutorial}.

Bayesian optimization works by assuming the objective function is drawn from a Gaussian process prior, i.e., $\mathcal{G}(\boldsymbol{\lambda}) \sim \mathcal{N}(\boldsymbol{0}, \boldsymbol{K})$, where $\boldsymbol{K}$ is the kernel matrix with entries $\boldsymbol{K}_{ij}=k(\boldsymbol{\lambda}_i, \boldsymbol{\lambda}_j)$. Given observation $\{\boldsymbol{\lambda}_n, \boldsymbol{g}_n\}_{n=1}^N$ from previous iterations, the posterior distribution of the function value at new point $\boldsymbol{\lambda}_{N+1}$ is also Gaussian:
$$ \mathcal{G}(\boldsymbol{\lambda}_{N+1}) \mid \{\boldsymbol{\lambda}_n, \boldsymbol{g}_n\}_{n=1}^N \sim \mathcal{N} (\mu (\boldsymbol{\lambda}_{N+1}), \sigma^2 (\boldsymbol{\lambda}_{N+1})) \eqno{(5)}$$
where
$$ \mu (\boldsymbol{\lambda}_{N+1}) = \boldsymbol{k}^{\top} \boldsymbol{K}^{-1} \boldsymbol{g}_{1:N}, \eqno{(6)}$$
$$ \sigma^2 (\boldsymbol{\lambda}_{N+1})) = k(\boldsymbol{\lambda}_{N+1}, \boldsymbol{\lambda}_{N+1}) - \boldsymbol{k}^{\top} \boldsymbol{K}^{-1} \boldsymbol{k}, \eqno{(7)} $$
$$ \boldsymbol{k} = \left[k(\boldsymbol{\lambda}_{N+1}, \boldsymbol{\lambda}_1), k(\boldsymbol{\lambda}_{N+1}, \boldsymbol{\lambda}_2), \cdots, k(\boldsymbol{\lambda}_{N+1}, \boldsymbol{\lambda}_n)  \right]^{\top}. \eqno{(8)}$$
Then the next location $\boldsymbol{\lambda}_{N+1}$ is determined by maximizing an acquisition function $\mathcal{A}$:
$$\boldsymbol{\lambda}_{N+1}=\arg \max _{\boldsymbol{\lambda} \in \boldsymbol{\Lambda}} \mathcal{A}(\boldsymbol{\lambda} \mid\left\{\boldsymbol{\lambda}_n, \boldsymbol{g}_n\right\}_{n=1}^N). \eqno{(9)}$$

In practice, we follow the recommendation in \cite{snoek2012practical} to use the automatic relevance determination Matérn 5/2 kernel as our kernel function and Expected Improvement as our acquisition function. We also benefit from the practical Bayesian optimization techniques proposed in \cite{snoek2012practical} to parallelize our Bayesian optimization procedures, thereby enhancing the optimization speed. In this work, we conduct two case studies by defining different performance metrics $\boldsymbol{g}$. In the first case, we measure the MSE between the outputs of the teacher and the student:
$$\boldsymbol{g} = \frac{1}{M}\sum^{M}_{m=1} \left\| \mathcal{F}_{teacher}(\boldsymbol{x}^O_m, \boldsymbol{\epsilon}) - \mathcal{F}_{student}(\boldsymbol{x}^O_m, \boldsymbol{\epsilon})\right\|_2^2,  \eqno{(10)} $$
where $\boldsymbol{\epsilon}$ is the noise sampled from the standard normal distribution and $M$ is the total number of validation data. Note that this is not the prediction error compared to the human motion ground truth. It is the discrepencies of outputs from the teacher model and the student model, and in this case, Bayesian optimization will identify a student model that performs similarly to the teacher model. 

For the second case, we aim to strike a balance between prediction accuracy and inference efficiency. For simplicity reason, we denote $\mathcal{F}_{teacher}(\boldsymbol{x}^O_m, \boldsymbol{\epsilon})$ as $\mathcal{F}_t$ and $\mathcal{F}_{student}(\boldsymbol{x}^O_m, \boldsymbol{\epsilon})$ as $\mathcal{F}_s$. After training a model, we calculate several unitless metrics:
$$ Ratio_{err} = \frac{1}{M} \sum^{M}_{m=1} \frac{\sqrt{\sum_{i}(\mathcal{F}_t{}_m^i - \mathcal{F}_s{}_m^i)^2}}{\sqrt{\sum_{i}(\mathcal{F}_t{}_m^i)^2}}, \eqno{(11)}$$
$$ Ratio_{acc} = \frac{Acc_{student} - Acc_{teacher}}{Acc_{teacher}}, \eqno{(12)}$$
$$ Ratio_{inf} = \frac{Time_{student} - Time_{teacher}}{Time_{teacher}}, \eqno{(13)}$$
where $i$ is the index of elements for the outputs from both the teacher model and the student model, $Acc$ represents the prediction accuracy in comparison to the human motion ground truth in the time domain, and $Time$ denotes the average inference time. Then the evaluation metric for the second case can be expressed as:
$$ \boldsymbol{g} = a \cdot Ratio_{err} + b \cdot Ratio_{acc} + c \cdot Ratio_{inf}, \eqno{(14)}$$
where $a$, $b$ and $c$ are weighting parameters.

Though we only demonstrate two different cases, it should be noted that one can always obtain models with varying performance by customizing different evaluation metrics.

\section{Experiments}

\begin{table*}[htbp]
\caption{Bayesian optimization results}
\begin{center}
\resizebox{2.0\columnwidth}{!}{%
\begin{threeparttable}
\begin{tabular}{ccccc|cccc}
\toprule
& \multicolumn{4}{c|}{Human3.6M} & \multicolumn{4}{c}{AMASS}\\ \cmidrule{2-9} 
Model & Learning Rate & \#Layers & Layer Dimension & \#Parameters & Learning Rate & \#Layers & Layer Dimension & \#Parameters \\ \midrule
One-step TransFusion & 0.0003 & 9 & 512 & 19.73M & 0.00015 & 13 & 512 & 28.30M \\ \midrule
SwiftDiff & 0.000674008545835 & 12 & 768 & 15.04M & 0.00068807828136 & 16 & 896 & 27.04M \\
SwiftDiff-balance & 0.000664587459924 & 10 & 768 & 12.67M & 0.000834066151193 & 10 & 896 & 17.37M \\ \bottomrule
\end{tabular}
\begin{tablenotes}
     \item[*] The one-step TransFusion model is not related to Bayesian optimization but is listed in the table for neural network architecture comparison.
\end{tablenotes}
\end{threeparttable}
}
\end{center}
\label{params}
\vspace{-0.2in}
\end{table*}

\begin{table*}[htbp]
\caption{Quantitative results with best-of-many, median-of-many and worst-of-many strategies on Human3.6M}
\centering
\resizebox{2.0\columnwidth}{!}{%
\begin{threeparttable}
\begin{tabular}{ccccccc}
\toprule
& \multicolumn{6}{c}{Human3.6M (Best-of-many / Median-of-many / Worst-of-many)} \\ \cmidrule{2-7} 
Model & Inference Time (sec) & APD (m) & ADE-B/M/W (m) & FDE-B/M/W (m) & MMADE-B/M/W (m) & MMFDE-B/M/W (m) \\ \midrule
DLow \cite{yuan2020dlow} & 0.0485 & 11.741 & 0.425 / 0.896 / 1.763 & 0.518 / 1.285 / 2.655 & 0.495 / 0.948 / 1.804 & 0.531 / 1.290 / 2.657 \\
MotionDiff \cite{wei2023human} & 1.010 & \textbf{15.353} & 0.411 / ------ / ------ & 0.509 / ------ / ------ & 0.508 / ------ / ------ & 0.536 / ------ / ------ \\
BeLFusion \cite{barquero2023belfusion} & 0.637 & 7.602 & 0.372 / 0.673 / 1.355 & 0.474 / 0.976 / 2.038 & \textbf{0.473} / 0.767 / 1.418 & \textbf{0.507} / 1.009 / 2.046 \\
HumanMAC \cite{chen2023humanmac} & 1.202 & 6.301 & 0.369 / 0.585 / 1.085 & 0.480 / 0.911 / 1.843 & 0.509 / 0.736 / 1.205 & 0.545 / 0.977 / 1.877 \\
TransFusion \cite{tian2024transfusion} & 1.122 & 5.975 & 0.358 / 0.575 / \textbf{1.063} & 0.468 / 0.898 / \textbf{1.758} & 0.506 / 0.729 / \textbf{1.179} & 0.539 / 0.967 / \textbf{1.791} \\ \midrule
Teacher Model & 0.215 & 6.216 & \textbf{0.355} / 0.582 / 1.156 & \textbf{0.462} / 0.903 / 1.884 & 0.502 / 0.735 / 1.263 & 0.533 / 0.973 / 1.913 \\
One-step TransFusion & 0.0113 & 5.988 & \textbf{0.355} / 0.575 / 1.130 & 0.465 / 0.890 / 1.844 & 0.502 / 0.729 / 1.240 & 0.534 / 0.961 / 1.874 \\
SwiftDiff & 0.0109 & 5.692 & 0.360 / 0.567 / 1.110 & 0.471 / \textbf{0.878} / 1.826 & 0.504 / 0.722 / 1.223 & 0.539 / \textbf{0.950} / 1.861 \\
SwiftDiff-balance & \textbf{0.00937} & 5.638 & 0.359 / \textbf{0.565} / 1.103 & 0.473 / 0.879 / 1.817 & 0.504 / \textbf{0.721} / 1.217 & 0.541 / \textbf{0.950} / 1.850 \\ \midrule
Percentage of Improvement & 80.680\% & -60.998\% & 0.838\% / 1.739\% / -3.763\% & 0.641\% / 2.227\% / -3.356\% & -6.131 \% / 1.097\% / -3.223\% & -5.325\% / 1.758\% / -3.294\% \\
\bottomrule
\end{tabular}
\begin{tablenotes}
     \item[*] Quantitative results for the baselines are taken from the original papers or calculated from publicly available pretrained models. The symbol '------' indicates that certain results are not reported in the baselines and that the pretrained model is not published online. Best results for each metric are highlighted in bold. The percentage of improvement is calculated by comparing the best results from the distilled models with the best results from previous works. A positive number indicates improvement.
\end{tablenotes}
\label{h36m}
\end{threeparttable}
}
\vspace{-0.1in}
\end{table*}

\begin{table*}[htbp]
\caption{Quantitative results with best-of-many, median-of-many and worst-of-many strategies on AMASS}
\centering
\resizebox{2.0\columnwidth}{!}{%
\begin{threeparttable}
\begin{tabular}{ccccccc}
\toprule
& \multicolumn{6}{c}{AMASS (Best-of-many / Median-of-many / Worst-of-many)} \\ \cmidrule{2-7} 
Model & Inference Time (sec) & APD (m) & ADE-B/M/W (m) & FDE-B/M/W (m) & MMADE-B/M/W (m) & MMFDE-B/M/W (m) \\ \midrule
DLow \cite{yuan2020dlow} & 0.0566 & \textbf{13.170} & 0.590 / 0.977 / 2.138 & 0.612 / 1.186 / 2.994 & 0.618 / 0.996 / 2.156 & 0.617 / 1.181 / 2.991 \\
BeLFusion \cite{barquero2023belfusion} & 0.711 & 9.376 & 0.513 / 0.817 / 1.791 & 0.560 / 1.069 / 2.237 & \textbf{0.569} / 0.857 / 1.815 & \textbf{0.591} / 1.074 / 2.236 \\
HumanMAC \cite{chen2023humanmac} & ------ & 9.321 & 0.511 / ------ / ------ & \textbf{0.554} / ------ / ------ & 0.593 / ------ / ------ & \textbf{0.591} / ------ / ------ \\
TransFusion \cite{tian2024transfusion} & 1.474 & 8.853 & \textbf{0.508} / 0.758 / 1.339 & 0.568 / 1.060 / 2.063 & 0.589 / 0.832 / 1.389 & 0.606 / 1.080 / 2.067 \\ \midrule
Teacher Model & 0.742 & 9.055 & \textbf{0.508} / 0.762 / 1.405 & 0.567 / 1.066 / 2.144 & 0.588 / 0.837 / 1.454 & 0.606 / 1.086 / 2.148 \\
One-step TransFusion & 0.0154 & 8.701 & 0.512 / 0.753 / 1.368 & 0.572 / 1.045 / 2.070 & 0.589 / 0.826 / 1.417 & 0.607 / 1.066 / 2.075 \\
SwiftDiff & 0.0140  & 8.411 & 0.515 / 0.744 / 1.360  & 0.579 / 1.033 / 2.041  & 0.590 / 0.818 / 1.409  & 0.611 / 1.055 / 2.044  \\
SwiftDiff-balance & \textbf{0.0116} & 8.145 & 0.515 / \textbf{0.737} / \textbf{1.330} & 0.579 / \textbf{1.028} / \textbf{2.028} & 0.589 / \textbf{0.810} / \textbf{1.378}  & 0.611 / \textbf{1.049} / \textbf{2.031}  \\ \midrule
Percentage of Improvement & 79.505\% & -33.933\% & -0.787\% / 2.770\% / 0.672\% & -3.249\% / 3.019\% / 1.697\% & -3.515 \% / 2.644\% / 0.792\% & -2.707\% / 2.328\% / 1.742\% \\ \bottomrule
\end{tabular}
\begin{tablenotes}
     \item[*] Quantitative results for the baselines are taken from the original papers or calculated from publicly available pretrained models. The symbol '------' indicates that certain results are not reported in the baselines and that the pretrained model is not published online. Best results for each metric are highlighted in bold. The percentage of improvement is calculated by comparing the best results from the distilled models with the best results from previous works. A positive number indicates improvement.
\end{tablenotes}
\label{amass}
\end{threeparttable}
}
\vspace{-0.1in}
\end{table*}

\subsection{Experimental Setup}

\begin{figure}
    \begin{center}
    \includegraphics[width=0.9\columnwidth]{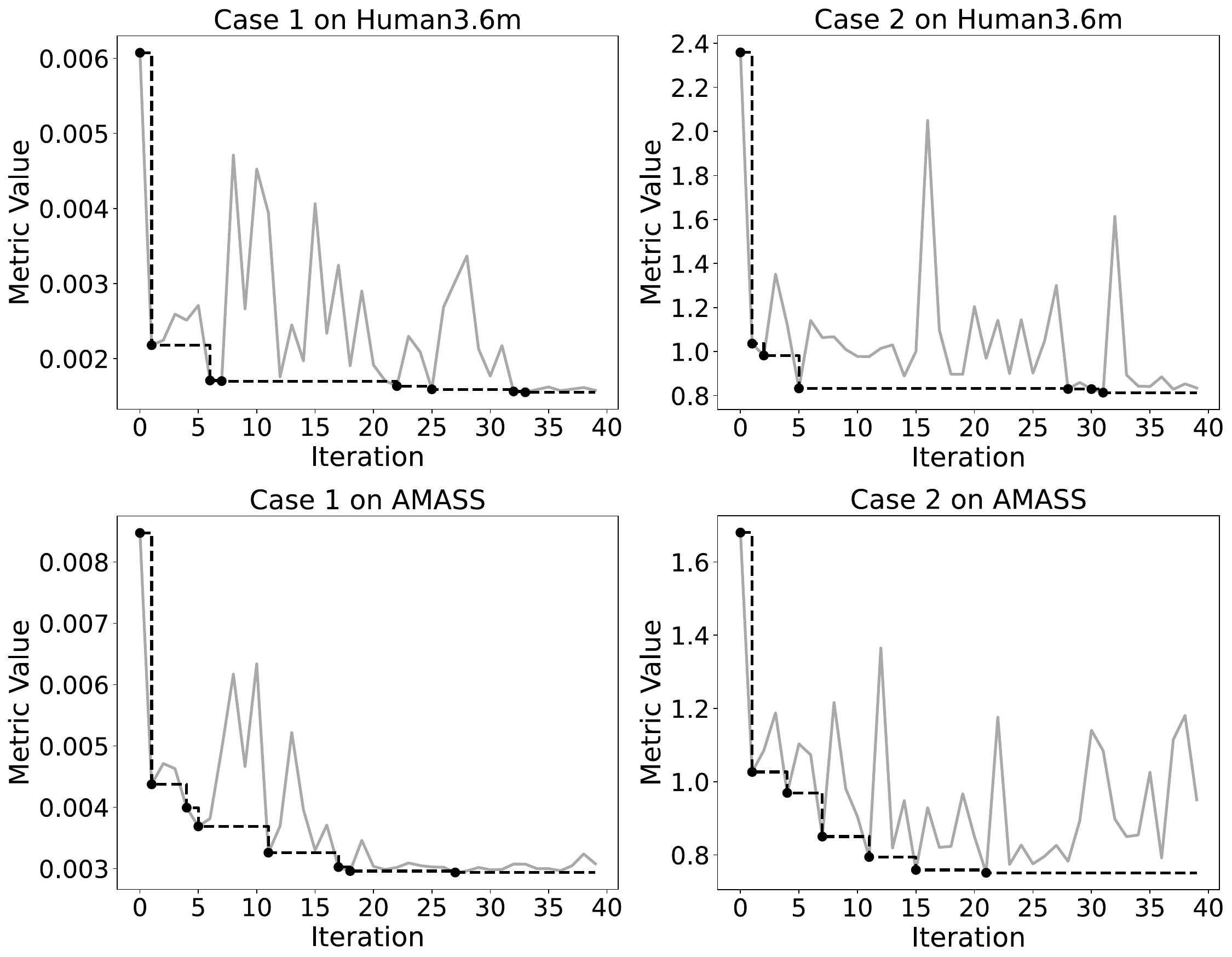}
    \caption{The progress of Bayesian optimization for both cases on both datasets.}
    \label{bo}
    \end{center}
    \vspace{-0.2in}
\end{figure}

\begin{figure*}
    \begin{center}
    \includegraphics[width=1.9\columnwidth]{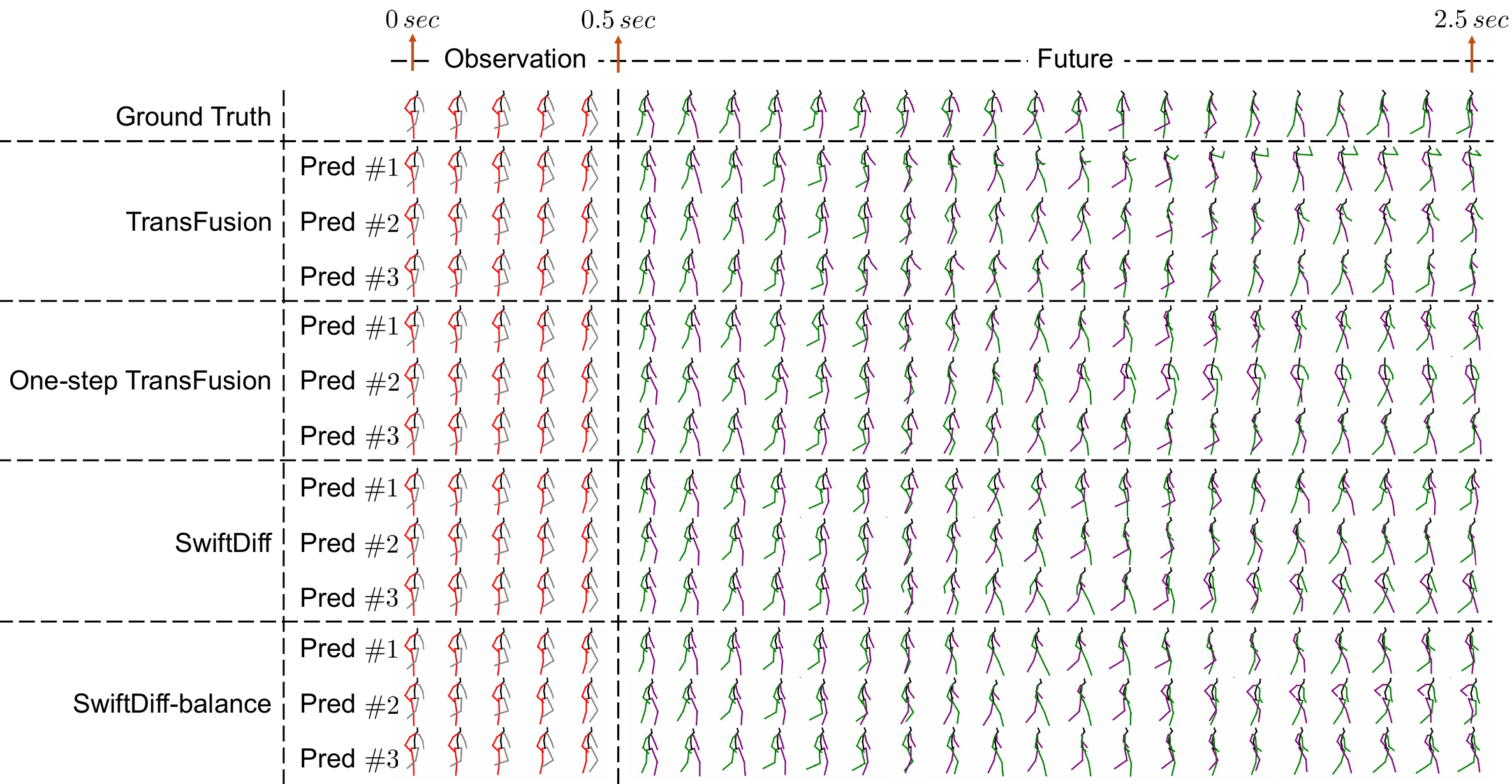}
    \caption{Visualization of predictions. Three predicted samples are displayed for each model. Variations are exhibited in the walking motion, and all predictions are semantically consistent with the historical motion. No detectable degradation is observed in the prediction results of the distilled models. More animations can be found at \textcolor{blue}{\url{https://github.com/sibotian96/SwiftDiff}}.}
    \label{visualization}
    \end{center}
    \vspace{-0.2in}
\end{figure*}

\textbf{Datasets:} We evaluate the performance of our work on two well-known benchmark datasets: Human3.6M \cite{ionescu2013human3} and AMASS \cite{mahmood2019amass}. Human3.6M is a widely used benchmark dataset for human motion prediction, containing 3.6 million frames of human poses captured at 50 Hz. To ensure a fair comparison with other studies, we follow the widely-used experimental pipeline, such as data processing and train-test split, proposed by \cite{yuan2020dlow}. Specifically, 25 frames (0.5 seconds) are used as the observation to forecast the following 100 frames (2 seconds). AMASS is a large-scale dataset that integrates 24 diverse datasets, all standardized to a common joint configuration. It has 9 million frames of data when downsampled to 60 Hz. Similarly, we follow the standard setting proposed in \cite{barquero2023belfusion} to make the comparison fair enough, and use 30 frames (0.5 seconds) to predict the subsequent 120 frames (2 seconds).

\textbf{Evaluation metrics:} We adopt the evaluation metrics proposed in \cite{tian2024transfusion} to validate the performance of our method: (1) Average Pairwise Distance (APD): APD calculates the mean L2 distance between all pairs of motion samples, providing a diversity measure among of predicted samples. The larger the number is, the more diverse the predictions are. (2) Average Displacement Error (ADE): ADE evaluates the prediction accuracy by calculating the average L2 distance over all time steps between the ground truth and predicted samples, and picks the one with smallest error for the best case. (3) Final Displacement Error (FDE): FDE is also a accuracy metric. It measures the minimum L2 distance in the final time frame between the prediction results and ground truth. (4) Multi-Modal ADE (MMADE): MMADE is the multi-modal version of ADE which evaluates the ability to capture the multi-modal nature of human motion. (5) Multi-Moddal FDE (MMFDE): Similarly, MMFDE is the multi-modal version of FDE.

Besides evaluating the accuracy of the best case, where only the closest sample to the ground truth among all prediction results is considered, the following metrics assess the median case and the worst case of all prediction samples, providing valuable insights into the overall prediction accuracy: (6) ADE-M, (7) FDE-M, (8) MMADE-M, and (9) MMFDE-M are metrics for median-of-many evaluation, while (10) ADE-W, (11) FDE-W, (12) MMADE-W, and (13) MMFDE-W are metrics for worst-of-many evaluation.

We also assess the inference time by providing the model with a single data point and measuring the time needed to predict one future motion. This setup is designed to reflect real-world scenarios, where only one observation is available at a time, and multiple predictions can be generated simultaneously in parallel. The average inference time from multiple experiments is taken as result.

\textbf{Baselines:} We perform a comparative evaluation against several state-of-the-art diffusion-based motion prediction works, which include MotionDiff \cite{wei2023human}, BeLFusion \cite{barquero2023belfusion}, HumanMAC \cite{chen2023humanmac} and TransFusion \cite{tian2024transfusion}. Additionally, we include a well-known VAE-based stochastic motion predictor DLow \cite{yuan2020dlow} as part of the baselines to offer a benchmark and give a sense of what performance in terms of accuracy is good.

\textbf{Implementation details:} The originial TransFusion uses 100 denoising steps. In this work, for the first distillation stage, we use TransFusion with 20 denoising steps for Human3.6M and 50 denoising steps for AMASS as teacher models, as they also exhibit state-of-the-art performance but is faster than the original model. We then train the one-step TransFusion model for 2,000 epochs on Human3.6M with a learning rate set to $3 \times 10^{-4}$, and for 4,000 epochs on AMASS with a learning rate set to $1.5 \times 10^{-4}$. For the second stage of distillation, the number of training epochs is set to 4,000 for both datasets. We utilize Bayesian optimization to tune the learning rate, number of layers, and layer dimension of SwiftDiff. The range of the learning rate is set to $\left[1\times10^{-4}, 1\times10^{-3}\right]$ for Human3.6M, and $\left[1\times10^{-4}, 1.5\times10^{-3}\right]$ for AMASS. The range of the number of layers is set to $\left[6,12\right]$ for Human3.6M, and $\left[10, 16\right]$ for AMASS. And the range of the layer dimension is set to $\left[256,768\right]$ for Human3.6M, and $\left[384, 896\right]$ for AMASS. For all experiments, we sample 50,000 data points from the training dataset in each epoch to train the model on both datasets and the batch size is set to 256. We use AdamW as the optimizer with a cosine annealing learning rate schedule and warm up the training process with 10\% of the total number of epochs. We run 40 iterations of Bayesian optimization with 5 objective function evaluations running in parallel for each case. We use ADE in the best-of-many case as $Acc$ in Eq. 12, and we run each model ten times to calculate the average inference time, denoted as $Time$ in Eq. 13. $a$, $b$ and $c$ in Eq. 14 are set to 15, 15 and 1, respectively.  All experiments are conducted on NVIDIA A100 GPU, except for the inference time evaluation. We use NVIDIA GeForce RTX 3080, which is more commonly available in gaming desktops, for inference speed testing to determine if the model can perform real-time predictions in practical scenarios such as HRC.

\subsection{Comparison With the State-of-The-Arts}

The results of Bayesian optimization for both cases defined in the methodology section are presented in Table \ref{params}, and the progress of Bayesian optimization in finding the optimum for both cases is shown in Fig. \ref{bo}. We denote the first case, which focuses solely on mimicking the teacher model, as SwiftDiff, and the second case, which strikes a balance between prediction accuracy and inference efficiency, as SwiftDiff-balance. We then compare our model with existing works in the best-of-many, median-of-many, and worst-of-many cases. The quantitative results are presented in Table \ref{h36m} for the Human3.6M dataset and Table \ref{amass} for the AMASS dataset. Our proposed models are comparable to other state-of-the-art works in terms of accuracy metrics; however, we significantly reduce the inference time compared with other baselines. It is worth to note that, we are not only faster than previous diffusion-based methods, we are also faster than the VAE-based DLow. Compared with DLow, we improve the inference time by around 80\% for both datasets. And the percentage of improvement reaches to around 98\% for both datasets if only compared with the best diffusion-based method. Given that the time interval between consecutive frames is 0.02 seconds for the Human3.6M dataset and 0.0167 seconds for the AMASS dataset, our model is the first diffusion-based approach proven capable of real-time human motion prediction. Moreover, our diversity metric is not as high as other prediction models, because we do not include any diversity-prompting techniques during training, since we aim to maintain meaningful variation, as shown in Fig. \ref{visualization}, instead of blindly increasing the diversity which may cause unrealistic predictions.

\section{Conclusion}

This paper presents a one-step MLP-based diffusion model for real-time human motion prediction, trained using knowledge distillation and Bayesian optimization. A pretrained state-of-the-art diffusion-based motion predictor is employed as the teacher model, and a two-stage distillation strategy is used to gradually reduce the complexity of the distillation task and effectively preserve the learned knowledge of the pretrained model. In the first distillation stage, the pretrained multi-denoising-step diffusion model is used to distill a one-step diffusion model with the same neural network architecture. In the second stage, the obtained one-step diffusion model serves as the new teacher to distill an even smaller and faster one-step MLP-based diffusion model. Compared to prior diffusion-based motion prediction works, our model significantly reduces inference time, achieving real-time prediction without noticeable degradation in prediction accuracy.

\bibliographystyle{IEEEtran}
\bibliography{ref}{}

\end{document}